%% file: main.tex
\documentclass[10pt,twocolumn,letterpaper]{article}

\usepackage[pagenumbers]{wacv} 


\usepackage{url}

\usepackage{xspace}
\usepackage{dsfont}
\usepackage{afterpage}

\usepackage{enumitem}
\usepackage{booktabs}
\usepackage{arydshln}
\usepackage{caption} 
\usepackage{multirow} 
\usepackage{enumitem}
\usepackage{colortbl} 
\usepackage{bbm}
\usepackage{algorithm}
\usepackage{algpseudocode}                  
\usepackage{setspace}                       
\usepackage{lipsum} 
\usepackage{subcaption} 
\usepackage{xcolor}

\usepackage{wrapfig}
\usepackage{pifont}

\usepackage{mathtools} 
\usepackage{bm} 
\usepackage{soul} 
\usepackage{nicefrac}
\usepackage{csquotes}

\definecolor{wacvblue}{rgb}{0.21,0.49,0.74}
\usepackage[pagebackref,breaklinks,colorlinks,allcolors=wacvblue]{hyperref}

\makeatletter
\newcommand*{\myfnsymbol}[1]{%
  \ensuremath{%
    \ifcase#1\or
      \text{\ding{41}}\or
      1\or
      2\or
      3\or
      4\or
      5\or
      6\or
      7\else
      8\fi
  }%
}
\def\@fnsymbol#1{\myfnsymbol{#1}}
\def\thempfootnote{\myfnsymbol{\c@mpfootnote}}
\makeatother

\def\wacvPaperID{2159} 
\def\confName{WACV}
\def\confYear{2027}

\title{From SRA to Self-Flow: Data Augmentation or Self-Supervision?}

\author{Dengyang Jiang$^{1}$ \hspace{10mm} Mengmeng Wang$^{2}$\thanks{Corresponding author. \qquad $^{\dagger}$Project leader.} \hspace{10mm} Harry Yang$^{1}$ \hspace{10mm} Jingdong Wang$^{3\dagger}$  \\ ~ \hspace{6mm}
\fontsize{10pt}{9.5pt}\selectfont
\hspace{-6mm} $^{1}$The Hong Kong University of Science and Technology \hspace{1mm} $^{2}$Zhejiang University of Technology \hspace{1mm} $^{3}$Baidu Inc.\\
}

\begin{document}
\maketitle

\begin{abstract}
Representation alignment has become an effective way to accelerate diffusion transformer training and improve generation quality. Recent self-alignment methods, such as SRA and Self-Flow, further remove the dependency on external pretrained encoders by constructing alignment within the diffusion model itself. However, the mechanism behind the improvement from SRA to Self-Flow, dual-time scheduling, remains under-examined: Self-Flow attributes its gain to interactions between tokens at different noise levels, where cleaner tokens help infer noisier ones. In this work, we revisit this explanation and ask whether the gain instead comes from data augmentation along the noise dimension. To disentangle these factors, we introduce Attention Separation, which preserves the same dual-timestep input as Self-Flow while blocking attention between tokens assigned to different noise levels. Surprisingly, removing such interaction does not degrade performance and can even improve it, suggesting that the improvement from SRA to Self-Flow mainly comes from data augmentation. Furthermore,We show that Attention Separation itself provides an augmentation effect by splitting a single image into multiple effective training parts to expand the training data. Based on these observations, we combine self-representation alignment with dual-timestep and attention-separation augmentation, and demonstrate the effectiveness of this design on ImageNet. Code: \url{https://github.com/vvvvvjdy/SRA/tree/main/SiT-SRA_DTS_AS}
\end{abstract}

\section{Introduction}
\label{introduction}
Enhancing the latent representation capability of Diffusion Transformers (DiTs)~\cite{dit,sit,sd3,z-image} during training has been demonstrated to accelerate convergence and improve generation quality~\cite{repa,repae,reg,sra,sra2,self-flow}. Prior works, such as REPA~\cite{repa}, attempt to achieve this by aligning the internal features of DiTs with those of a frozen, pre-trained image encoder (e.g., DINOv2~\cite{dinov2}). However, this external alignment strategy often falls short in scenarios where a sufficiently powerful encoder is absent, or when scaling up the training data and model size for DiTs~\cite{waver2025,self-flow}. To address this, recent research has pivoted toward  representation alignment within the DiT itself~\cite{sra,layersync,self-flow}. Pioneer work like Self-Representation Alignment (SRA)~\cite{sra}, which aligns latent representations in earlier layers under higher noise conditions with those in deeper layers under lower noise levels of the same model to progressively reinforcing internal representation learning. Subsequently, Self-Flow~\cite{self-flow} extends this self-representation alignment paradigm to multi-modal scenarios and larger scales (e.g, Text-to-Image, Text-to-Video, Text-to-Audio), demonstrating that it consistently outperforms the external alignment methods like REPA, and the self-representation alignment baseline SRA. 

\begin{figure}[t]
    \centering
    \includegraphics[width=1\linewidth]{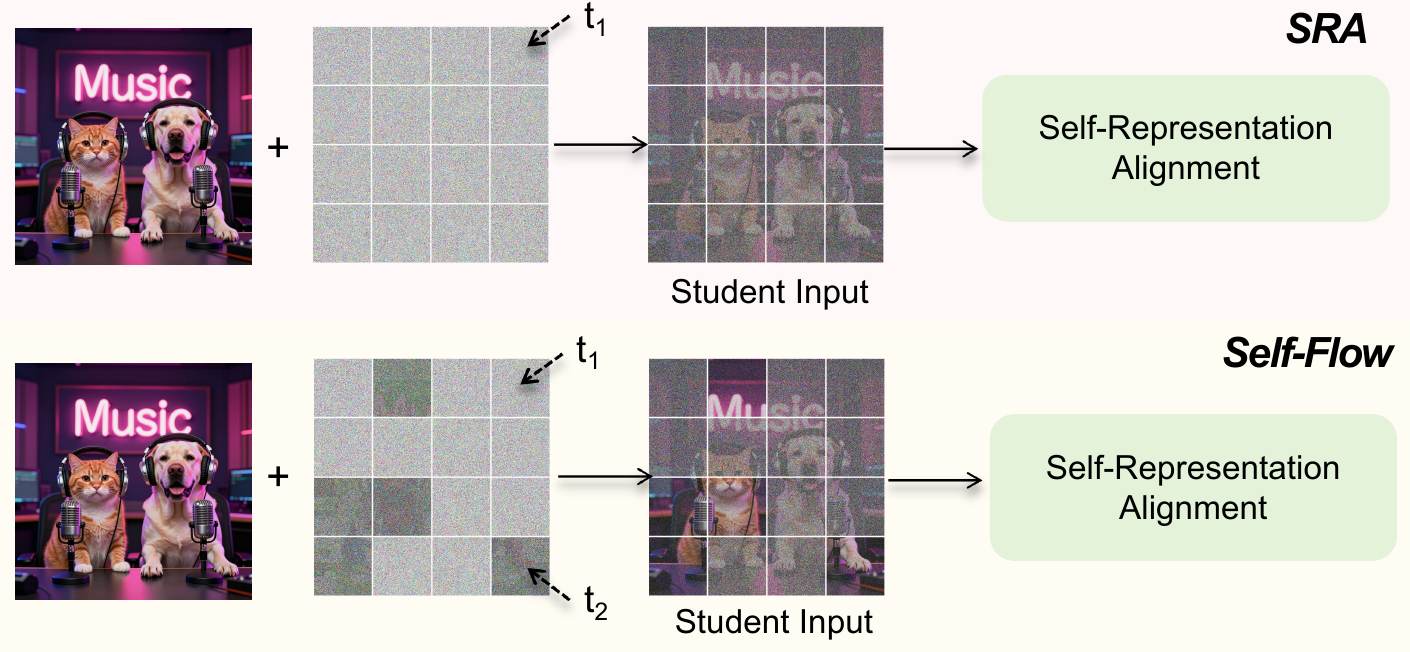}
    \caption{\textbf{Difference Between Self-Flow and SRA.} Self-Flow adopts SRA's Self-Representation Alignment method while differs in  the student input processing: SRA utilizes a single noise level ($t_1$) for all student input tokens, whereas Self-Flow employs a dual-timestep scheduler where tokens at two distinct noise levels ($t_1$ and $t_2$) coexist in the same image.}
    \label{fig:diff}
\end{figure}

Notably, as illustrated in Figure~\ref{fig:diff}, Self-Flow also adopts the Self-Representation Alignment method pioneered by SRA. The key distinction, however, lies in how the input samples are processed for the student. Specifically, Self-Flow introduces a dual-timestep scheduling, where a single input sample to the student model contains patches corrupted by two distinct noise-levels. Consequently, the performance gains achieved by Self-Flow over SRA are primarily attributed to this specific design. In Self-Flow paper, the explanation of the mechanism of this dual-timestep scheduling is: \textit{"by applying different noise levels to different tokens, the model is encouraged to use cleaner tokens to infer noisy tokens. This drives learning strong representations alongside generative capabilities."} Nevertheless, we question that dose these improvements indeed stem from superior self-supervision achieved by interactions of different noise-level tokens? 

In this work, we revisit the mechanism behind the gains of dual-timestep scheduling. Rather than attributing the improvement solely to better self-supervision by interactions, \textit{we argue that this design also functions as a form of data augmentation for diffusion training}. Here, data augmentation does not directly alter the semantic content of the clean image\cite{cutmix,mixup}; instead, it expands the effective training distribution along the noise dimension. By assigning different noise level to different token subsets, a single clean sample is presented to the model under more diverse noise states, allowing the model to observe more noise-conditioned variants of the same data within training, thus expands the effective training data for the model.

\begin{figure}[t]
\centering
\includegraphics[width=1\linewidth]{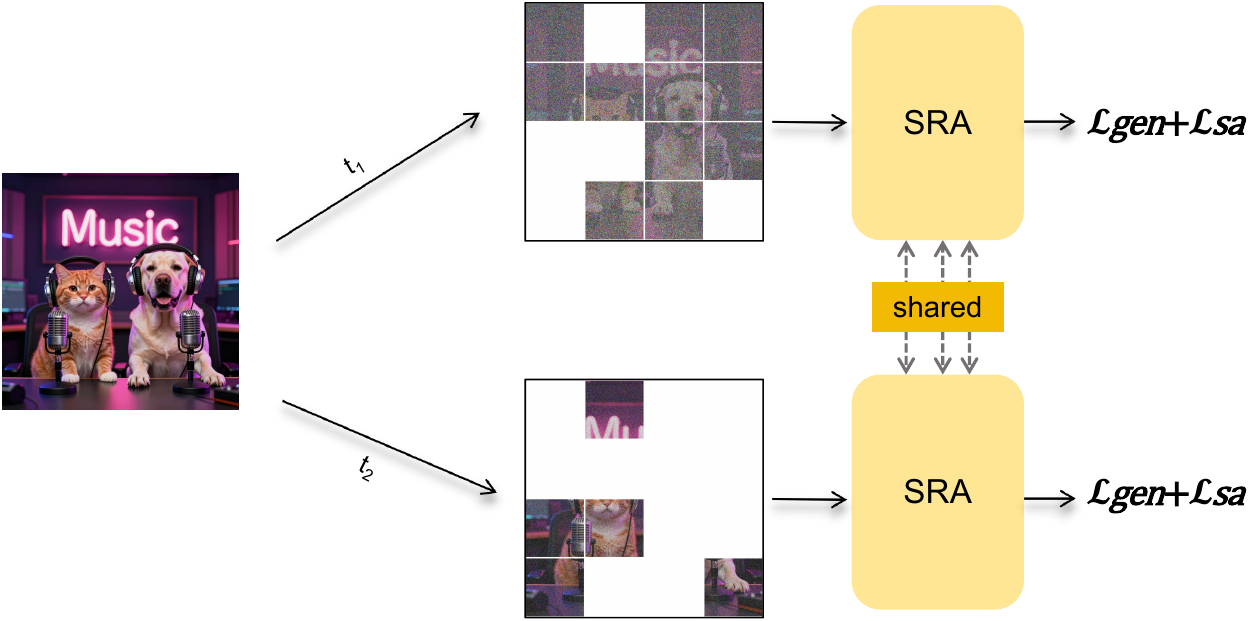}
\caption{
\textbf{Attention Separation disentangles self-supervision from augmentation.}
Given a sample with dual-timestep scheduling, Attention Separation preserves the heterogeneous-noise input but partitions attention into independent timestep groups, so tokens at $t_1$ cannot interact with tokens at $t_2$. This removes token interactions while keeping the noise-state augmentation introduced by dual-timestep scheduling. Meanwhile, Attention Separation can also be interpreted as creating multiple part-conditioned views of one image to expand the training distribution, thereby also acting as a data augmentation.
}
\label{fig:attention_separation_aug}
\end{figure}

To verify this hypothesis, we introduce Attention Separation, as illustrated in Figure~\ref{fig:attention_separation_aug}. The key idea is to preserve the same dual-timestep input as Self-Flow while removing the interaction between tokens at different noise levels. Specifically, tokens assigned to the same timestep can attend to each other, whereas tokens assigned to different timesteps are blocked from interacting. This creates a controlled setting: if the improvement of Self-Flow mainly comes from cleaner tokens guiding noisier tokens through attention, removing such interaction should degrade performance; if the gain remains, the dual-timestep scheduler is more likely acting as noise-state augmentation.

This observation further leads us to reinterpret Attention Separation itself as a form of data augmentation. When applied under single-timestep training, all tokens share the same noise level. Nevertheless, Attention Separation still improves training, we analyze that with such separation, each token group acts as a partial observation of the original image. These partial views are processed by the same shared-parameter model and optimized with the same denoising and self-alignment objectives in a single iteration as shown in Figure~\ref{fig:attention_separation_aug}. Thus, each image yields multiple effective training samples with different content subsets to expand the effective training distribution.

We evaluate this interpretation through controlled ablations and system-level comparisons, our final training scheme improves over previous self-alignment baselines on most metrics and remains on par with, or better than, the external-encoder baseline on ImageNet. 

In summary, our main contributions are as follows:
\begin{itemize}[leftmargin=*]
    \item We revisit the mechanism behind the improvement from SRA to Self-Flow and show that dual-timestep scheduling is better explained as data augmentation rather than self-supervision.
    \item We introduce Attention Separation, a controlled operation that blocks interactions between tokens at different noise levels, and further show that it can also serve as a data augmentation.
    \item We combine dual-timestep scheduling and Attention Separation within self-representation alignment, achieving stronger results than previous self-alignment baselines on most metrics and competitive performance with external-encoder alignment.
\end{itemize}

\section{Related Work}
\label{related-work}

\subsection{Representation Alignment for Generation}
Improved  latent representations of diffusion models can accelerate convergence and enhance generation~\cite{repa,repae,ddt,reg,sra,sra2,self-flow}.
One prominent avenue is leveraging discriminative priors from pretrained vision encoders for alignment~\cite{repa,repae,reg,irepa}.  REPA~\cite{repa} pioneered this paradigm by aligning intermediate diffusion features with representations from external visual encoders such as DINOv2~\cite{dinov2}. Building upon this paradigm of external representation alignment, subsequent studies have introduced refinements in alignment mechanisms~\cite{irepa,reg}, training strategies~\cite{repanotwork,repae}, etc.. Meanwhile, an alternative research avenue has emerged that dispenses with external encoders entirely, opting instead to perform representation alignment internally within the Diffusion Transformer~\cite{sra,sra2,self-flow,layersync}. SRA~\cite{sra} pioneered this paradigm by aligning latent representations in earlier layers of  with those in deeper layers of the same model. Following work Self-Flow~\cite{self-flow} extend this idea to larger-scale  settings and more modalities (e.g., text-to-image, text-to-video, and text-to-audio). And shows that self alignment paradigm consistently outperforms the external alignment counterpart. In this work, we revisit this internal alignment paradigm and study whether the improvement of dual-timestep scheduling comes from stronger self-supervision by interactions or from the augmented noise states introduced during training.

\subsection{Visual Self-Supervised Learning}
 Self-Supervised Learning (SSL) leverages pretext tasks to learn robust representations without manual labels~\cite{dino,moco,mae,jepa,ibot,cae,beit,byol}. For example, Masked Autoencoders (MAE)~\cite{mae} reconstruct masked image patches to provide strong downstream initializations. MoCo~\cite{moco} introduces a momentum encoder and a queue-based dictionary to learn instance-discriminative representations from augmented image views. DINO~\cite{dino} instead adopts a self-distillation framework, where a student network learns to match a momentum teacher across different views without using negative samples. Follow-up works such as DINOv2~\cite{dinov2} further extend this idea to large-scale settings, producing strong general-purpose visual representations. In visual generation, Self-Supervised Learning  is also applied in the training process like pre-training~\cite{maskdit,sddit,sra,self-flow} and post-training~\cite{dopsd}.  In this work, we further examine this self-supervised interpretation in diffusion training, showing that the benefit of dual-timestep scheduling does not primarily rely on self-supervision, but can be explained by its data augmentation effect.

\subsection{Data Augmentation}
Data augmentation enlarges the effective training data by constructing task-preserving variants of existing samples. In visual representation learning, classical strategies include Mixup~\cite{mixup}, which interpolates images and labels, Manifold Mixup~\cite{manifoldmixup}, which performs interpolation in hidden spaces, and CutMix~\cite{cutmix}, which replaces image regions across samples. In self-supervised learning, augmentations also define different views for representation learning, as in SimCLR~\cite{simclr}. For diffusion-based generation, data augmentation has recently been explored to improve both discriminative and generative training. Diffusion-generated synthetic images can improve ImageNet classification~\cite{syntheticdiffusion,lbgen}, while Degeorge et al.~\cite{imagenet_t2i} show that competitive text-to-image diffusion models can be trained from ImageNet alone with synthetic long captions and image augmentations such as CutMix and crop-based training. In this work, we focus on understanding the mechanism behind the gains 
of  dual-timestep scheduling~\cite{self-flow}, and show through Attention Separation that it is better viewed as data augmentation.

\section{Preliminary: SRA and Self-Flow}

Since both SRA and Self-Flow are built upon the Flow Matching Models~\cite{sit,sd3}, we begin by introducing the standard training objective of Flow Matching models. Subsequently, we elaborate on the formulations of SRA and Self-Flow, followed by an  analysis of the key distinctions between these two methods.

\paragraph{Flow matching.}
We consider a conditional flow-matching DiT parameterized by $\theta$. Given a clean sample $x_0 \sim p_{\mathrm{data}}$, condition $c$, and Gaussian noise $x_1 \sim \mathcal{N}(0,I)$, a noisy sample at timestep $t \in [0,1]$ is obtained by the linear path~\cite{flow-matching,rectified-flow}:
\begin{equation}
    x_t = (1-t)x_0 + t x_1 ,
\end{equation}
where a larger $t$ corresponds to a higher noise level. The model predicts the velocity field along this path and is trained with the standard generation objective:
\begin{equation}
    \mathcal{L}_{\mathrm{gen}}
    =
    \mathbb{E}_{x_0,x_1,t,c}
    \left[
    \left\|
    v_\theta(x_t,t,c) - (x_1-x_0)
    \right\|_2^2
    \right].
\end{equation}

\paragraph{SRA.}
SRA~\cite{sra} introduces a self-alignment objective without relying on an external representation encoder. Let $h_\theta^{m}(\cdot)$ denote the feature map from the $m$-th layer of the student DiT, and let $h_{\bar{\theta}}^{n}(\cdot)$ denote the feature map from the $n$-th layer of its EMA teacher, where usually $m \leq n$. For a high-noise timestep $t$ and a lower-noise timestep $s<t$, SRA feeds $x_t$ into the student and $x_s$ into the EMA teacher. The early-layer student representation is then projected by a lightweight head $g_\psi$ and aligned to the stop-gradient teacher representation:
\begin{equation}
    \mathcal{L}_{\mathrm{SRA}}
    =
    \mathbb{E}
    \left[
    \frac{1}{N}
    \sum_{i=1}^{N}
    d\left(
    g_\psi\!\left(h_\theta^{m}(x_t,t,c)\right)_i,
    \mathrm{sg}\!\left[h_{\bar{\theta}}^{n}(x_s,s,c)_i\right]
    \right)
    \right],
\end{equation}
where $N$ is the number of tokens, $i$ indexes a token, and $d(\cdot,\cdot)$ is a feature distance such as cosine or $\ell_2$ distance. The overall objective is:
\begin{equation}
    \mathcal{L}=\mathcal{L}_{\mathrm{gen}}+\lambda \mathcal{L}_{\mathrm{SRA}} .
\end{equation}
Thus, SRA constructs self-supervision from two asymmetries: the student observes a noisier input than the teacher, and an earlier student layer is encouraged to match a later teacher layer.

\paragraph{Self-Flow.}
Self-Flow~\cite{self-flow} follows the same EMA-based self-alignment principle introduced in SRA, but changes how the student input is constructed. Instead of assigning a single timestep to all tokens, it samples two timesteps $t$ and $s$, defines $t_{\mathrm{hi}}=\max(t,s)$ and $t_{\mathrm{lo}}=\min(t,s)$, and builds a token-wise timestep vector $\boldsymbol{\tau}\in[0,1]^N$:
\begin{equation}
    \tau_i =
    \begin{cases}
    t_{\mathrm{hi}}, & i \in M,\\
    t_{\mathrm{lo}}, & i \notin M,
    \end{cases}
\end{equation}
where $M$ is a randomly sampled token mask. The student input is then mixed at the token level,
\begin{equation}
    x_{\boldsymbol{\tau},i}
    =
    (1-\tau_i)x_{0,i}+\tau_i x_{1,i},
\end{equation}
while the EMA teacher receives the  cleaner input $x_{t_{\mathrm{lo}}}$. Its representation objective can be written as:
\begin{equation}
    \mathcal{L}_{\mathrm{SF}}
    =
    \mathbb{E}
    \left[
    \frac{1}{N}
    \sum_{i=1}^{N}
    d\left(
    g_\psi\!\left(h_\theta^{m}(x_{\boldsymbol{\tau}},\boldsymbol{\tau},c)\right)_i,
    \mathrm{sg}\!\left[h_{\bar{\theta}}^{n}(x_{t_{\mathrm{lo}}},t_{\mathrm{lo}},c)_i\right]
    \right)
    \right],
\end{equation}
with $g_\psi$ set to the identity when no projection head is used.

This formulation highlights the essential difference between SRA and Self-Flow: SRA assigns a single noise level to the entire student input. In contrast, Self-Flow introduces dual-timestep scheduling, where tokens with different noise levels coexist within the same student input. Self-Flow attributes its improvement to interactions among tokens at different noise levels: cleaner tokens provide contextual cues for noisier tokens, thereby encouraging stronger self-supervised representation learning. However, this scheduler also changes the training data,  the student observes multiple noise levels within one sample instead of a single global timestep, which exposes the model to more diverse noise-level instances and can be viewed as token-wise data augmentation for the denoising task. Therefore, we argue that the gain of Self-Flow over SRA may stem from two entangled factors: interactions for better self-supervision, and heterogeneous-noise training as data augmentation.

\section{Isolate Effect by Attention Separation}

\begin{figure}[h]
    \centering
    \includegraphics[width=1\linewidth]{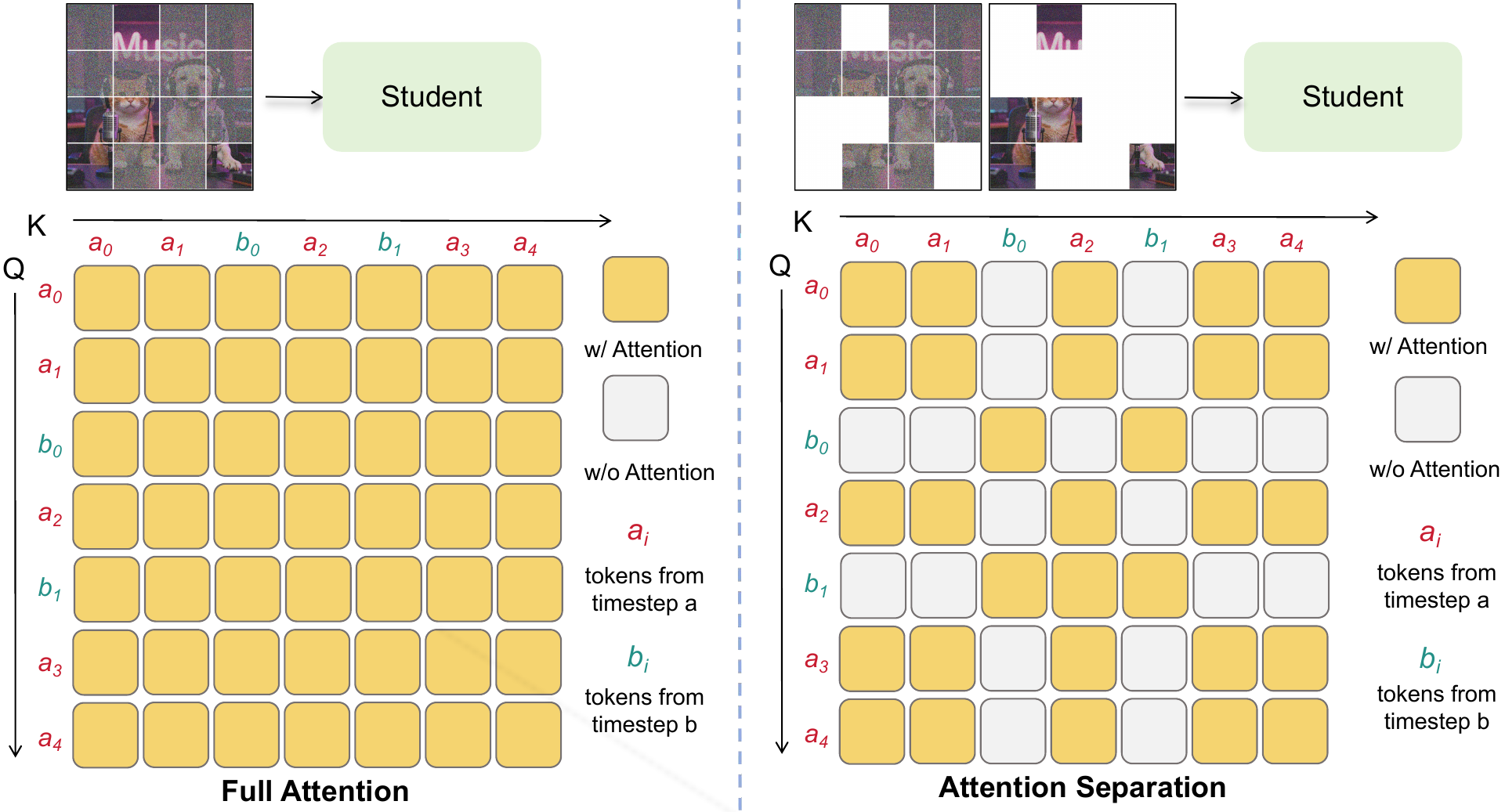}
    \caption{\textbf{Attention Separation Visualization.} 
Given a dual-timestep input, full attention allows all tokens to attend to each other regardless of their assigned timestep. In contrast, Attention Separation applies a block-diagonal attention mask: tokens from the same timestep group can interact, while tokens from different timestep groups are blocked.}
    \label{fig:att_s}
\end{figure}

To disentangle these two factors, we design an Attention Separation operation that preserves the dual-timestep noise  while selectively removes token interaction from different noise-levels.  Specifically, we keep the same dual-timestep scheduling as Self-Flow, which means that the model still observes heterogeneous noise levels within each training sample. The only modification is in the self-attention computation. Let $r_i=\mathbbm{1}[i\in M]$ be the group indicator of token $i$. We construct a binary attention mask:
\begin{equation}
    A^{\mathrm{sep}}_{ij}
    =
    \mathbbm{1}[r_i=r_j]
    =
    \mathbbm{1}[\tau_i=\tau_j],
\end{equation}
which allows attention only between tokens of the same noise level. For a self-attention layer with queries, keys, and values $(Q,K,V)$, Attention Separation computes:
\begin{equation}
    \mathrm{Attn}^{\mathrm{sep}}(Q,K,V)_i
    =
    \sum_{j=1}^{N}
    \frac{
    A^{\mathrm{sep}}_{ij}\exp(q_i^\top k_j/\sqrt{d})
    }{
    \sum_{l=1}^{N} A^{\mathrm{sep}}_{il}\exp(q_i^\top k_l/\sqrt{d})
    }
    v_j .
\end{equation}
Equivalently, tokens from different timestep groups are assigned $-\infty$ attention logits before the softmax. As shown in Figure~\ref{fig:att_s}, this turns the full attention matrix  into a block-diagonal one: tokens assigned to the same noise level can attend to each other, whereas tokens from different noise levels are prevented from interacting through attention. This controlled setting preserves the heterogeneous-noise training but removes interactions. Therefore, comparing Self-Flow with its attention-separated counterpart allows us to isolate whether the observed gain mainly comes from self-supervision by interaction or from the dual-timestep noise as data augmentation.

\section{Data Augmentation Matters More}

To answer the question raised above, we conduct controlled ablations on ImageNet $256\times256$~\cite{imagenet} using SiT-B~\cite{sit} by default. We report FID-10K~\cite{fid} and IS~\cite{IS} at different training iterations. Unless otherwise specified, the training and inference hyperparameters settings follow the default choices used in SRA and Self-Flow~\cite{sra,self-flow}. Our goal is to disentangle whether the gain mainly comes from token interaction or from heterogeneous-noise  data augmentation.

\paragraph{Removing interaction does not weaken dual-timestep training.}
 As Attention Separation blocks attention between tokens assigned to different noise levels while preserving the same heterogeneous noise assignment. If the gain mainly came from cleaner tokens guiding noisier tokens through self-attention, this intervention should degrade performance. Table~\ref{tab:dual_timestep_separation}  shows the ablation results of whether isolates the role of  token interaction of different noise-levels under dual-timestep scheduling. It can be seen that Attention Separation achieves comparable FID at 100K and improves both FID and IS at later stages, reducing FID from $25.19$ to $25.06$ and increasing IS from $66.75$ to $72.94$ at 800K. This result supports the interpretation that the dual-timestep benefit does not primarily rely on the interactions of tokens.

 \begin{table}[h]
\centering
\small
\setlength{\tabcolsep}{5pt}
\caption{\textbf{Ablation under dual-timestep scheduling}. Both rows use the same dual-timestep noise assignment; the only difference is whether tokens from different noise levels are allowed to interact through self-attention. The comparable or improved performance indicates that dual-timestep scheduling does not primarily rely on token interaction across noise levels.}
\label{tab:dual_timestep_separation}
\begin{tabular}{llccc}
\toprule
Attention & Metric & 100K & 400K & 800K \\
\midrule
\multirow{2}{*}{Full attention}
& FID $\downarrow$ & 58.16 & 30.20 & 25.19 \\
& IS $\uparrow$  & 23.43 & 54.44 & 66.75 \\
\midrule
\multirow{2}{*}{Attention Separation}
& FID $\downarrow$ & 58.57 & 29.89 & 25.06 \\
& IS $\uparrow$  & 25.20 & 58.29 & 72.94 \\
\bottomrule
\end{tabular}
\end{table}

\begin{figure}[h]
\centering
\includegraphics[width=1\linewidth]{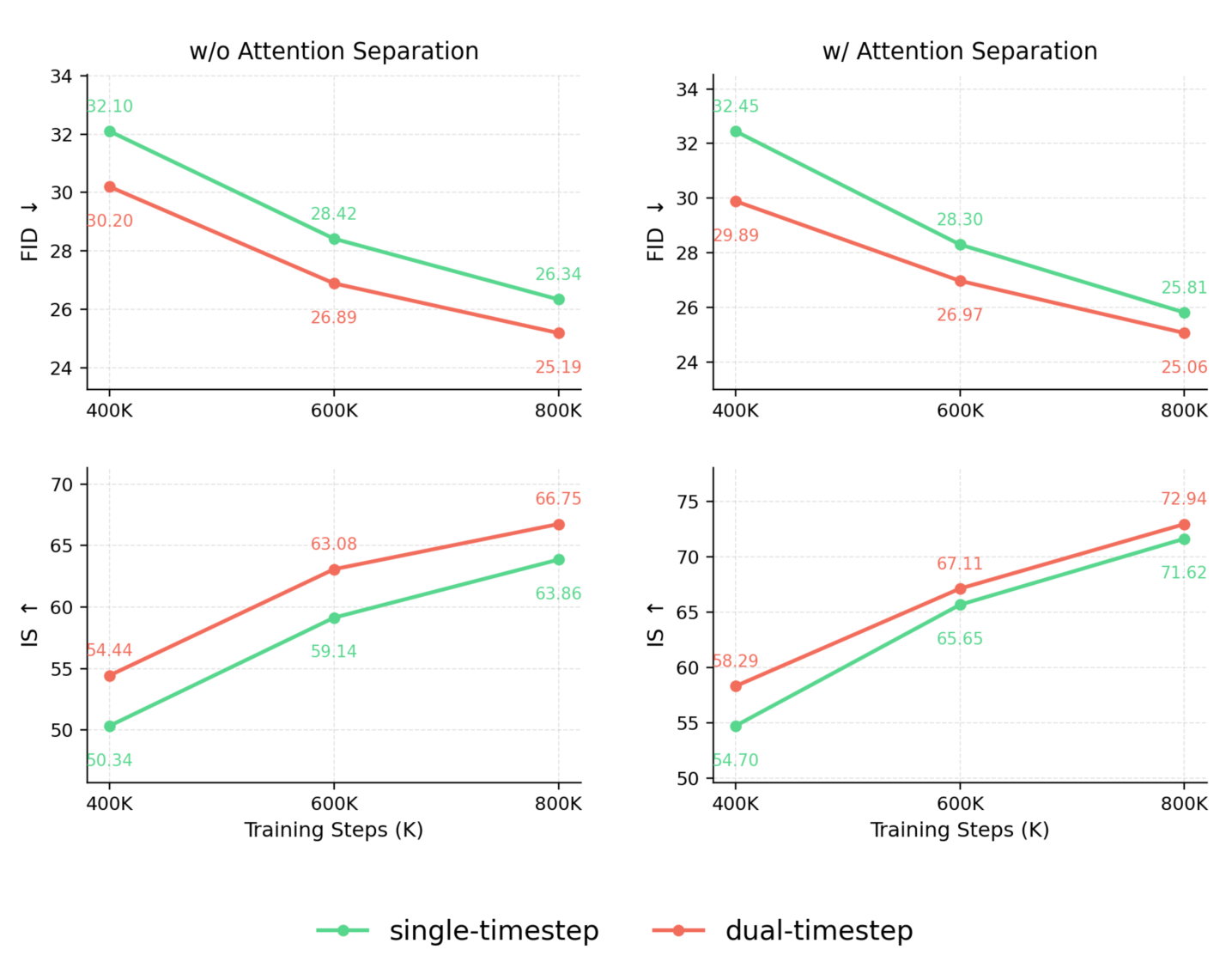}
\caption{Comparison between single-timestep and dual-timestep training under matched attention settings. \textbf{Left}: without Attention Separation. \textbf{Right}: with Attention Separation. Dual-timestep training consistently improves over single-timestep training under both full attention and attention-separation, supporting the view that its benefit comes from noise-state augmentation.}
\label{fig:single_dual_matched_attention}
\end{figure}

\paragraph{Dual-timestep scheduling primarily works as data augmentation.}
Figure~\ref{fig:single_dual_matched_attention} compares single-timestep and dual-timestep training under matched attention settings. With full attention, dual-timestep training consistently improves FID from $32.10/28.42/26.34$ to $30.20/26.89/25.19$ at 400K/600K/800K iterations. More importantly, the same improvement remains under Attention Separation, where interactions between different noise-level tokens are explicitly blocked: dual-timestep training improves FID from $32.45/28.30/25.81$ to $29.89/26.97/25.06$ and also improves IS across all training stages. This indicates that the gain does not rely on stronger self-supervision induced by token interactions. Instead, dual-timestep scheduling changes the training data seen by the student: each image is decomposed into token subsets observed at different noise levels, exposing the model to more noise-state variants within the same training iteration, thus expand the effective training distribution. Therefore, its effect is better understood as data augmentation to expose the model to more data.

\section{Attention Separation Is Also a Data Augmentation}

Table~\ref{tab:single_timestep_separation} compares full attention and Attention Separation under the single-timestep setting.  In this case, it do not has any interaction from tokens in different noise level, since all tokens share a timestep. It only partitions the image tokens into several non-interacting groups. However, We observe that Attention Separation still brings clear gains, especially in IS, even when all tokens share the same timestep. To investigate the source of the performance gains, we conducted the following equivalent substitution analysis.

As illustrated in Figure~\ref{fig:attention_separation_aug} and Figure~\ref{fig:att_s}, Attention Separation can be interpreted as converting one training image into multiple part-conditioned training views. Whether the two groups use different timesteps ($t_1 \neq t_2$) or the same timestep ($t_1=t_2$), the separation mask makes each token group behave like a partial observation of the original image. These partial views are processed by the same model with shared parameters and optimized with the same denoising and self-alignment objectives in one iteration. Equivalently, a single image provides multiple effective training samples, each containing a different subset of the full image. This expands the effective training distribution without introducing external data. Together with the results in Table~\ref{tab:single_timestep_separation}, indicating that Attention Separation itself also acts as data augmentation along the sample view, while dual-timestep scheduling augments the sample along the noise-state dimension.

\begin{table}[h]
\centering
\small
\setlength{\tabcolsep}{5pt}
\caption{\textbf{Effect of Attention Separation under single-timestep training}. Since all tokens share the same timestep, the separation mask only partitions image tokens into non-interacting parts. The gains under single-timestep training show that Attention Separation can be beneficial even without cross-noise tokens, suggesting a augmentation effect.}
\label{tab:single_timestep_separation}
\begin{tabular}{llccc}
\toprule
Attention & Metric & 100K & 400K & 800K \\
\midrule
\multirow{2}{*}{Full attention}
& FID $\downarrow$ & 62.95 & 32.10 & 26.34 \\
& IS $\uparrow$  & 22.26 & 50.34 & 63.86 \\
\midrule
\multirow{2}{*}{Attention Separation}
& FID $\downarrow$ & 62.39 & 32.45 & 25.81 \\
& IS $\uparrow$  & 24.08 & 54.70 & 71.62 \\
\bottomrule
\end{tabular}
\end{table}

\paragraph{Effect of the mask ratio.}
We further study how the mask ratio affects dual-timestep training with Attention Separation. Let $\alpha=|M|/N$ denote the fraction of tokens assigned to one timestep group; thus, $\alpha=0.25$ partitions an image into two groups with $25\%$ and $75\%$ tokens. Table~\ref{tab:mask_ratio_800k} reports the ablation results.

\begin{table}[h]
\centering
\small
\setlength{\tabcolsep}{5pt}
\caption{\textbf{Effect of mask ratio}. We compare single-timestep training, dual-timestep training with full attention, and dual-timestep training with Attention Separation in different mask ratio. All results are tested on the model trained with 800K iterations. A mild ratio preserves the augmentation benefit, while larger ratios hurts performance due to a stronger training--inference mismatch.}
\label{tab:mask_ratio_800k}
\begin{tabular}{lllcc}
\toprule
Time Scheduling & Attention & Mask Ratio & FID $\downarrow$ & IS $\uparrow$ \\
\midrule
Single & Full  & - & 26.34 & 63.86 \\
\midrule
\multirow{2}{*}{Dual}
& Full  & 0.25 & 25.19 & 66.75 \\
& Separation & 0.25 & 25.06 & 72.94 \\
\midrule
\multirow{2}{*}{Dual}
& Full  & 0.35 & 24.87 & 65.46 \\
& Separation & 0.35 & 27.50 & 68.12 \\
\midrule
\multirow{2}{*}{Dual}
& Full  & 0.50 & 24.39 & 66.62 \\
& Separation & 0.50 & 38.19 & 67.20 \\

\bottomrule
\end{tabular}
\end{table}

When full attention is used, changing the mask ratio does not harm dual-timestep training, and the performance remains consistently better than the single-timestep baseline. This is expected under our augmentation interpretation: regardless of the exact partition ratio, the model is still exposed to more noise states within each image, while full-image attention allows every token to access the complete image context. Thus, changing the mask ratio mainly changes the relative amount of tokens in one group, but does not prevent the model from learning with global spatial context. However, the behavior changes once Attention Separation is applied. While $\alpha=0.25$ achieves the best IS and comparable FID, larger ratios degrade FID substantially, especially at $\alpha=0.50$. We hypothesize that this degradation comes from a stronger training--inference mismatch induced by overly balanced separation. During training, Attention Separation decomposes each image into two non-interacting token groups, so each attention component can only aggregate information from a partial view of the image. As the mask ratio approaches $0.50$, both groups become incomplete views with similar size, and neither group consistently preserves most of the global image context. In contrast, inference uses standard full-image attention, where all tokens interact globally. This gap between part-level training and full-image inference becomes more severe at larger mask ratios, leading to the observed FID degradation.

\begin{figure}[h]
\centering
\includegraphics[width=1\linewidth]{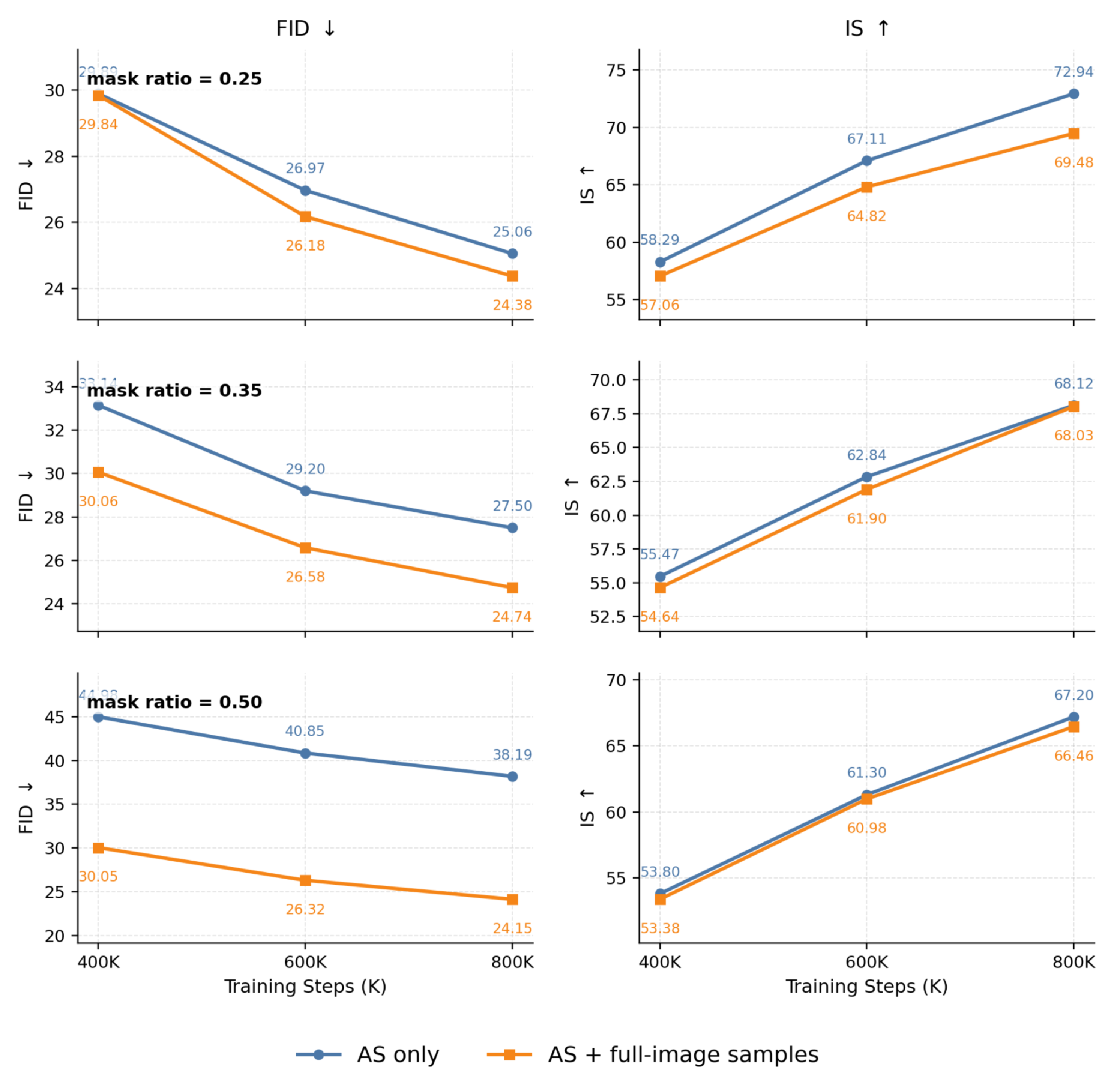}
\caption{\textbf{Effect of adding full-image samples}. We compare Attention Separation applied to all dual-timestep samples with a mixed setting that includes full-image single-timestep samples. Adding full-image samples substantially reduces the mismatch caused by strong separation.}
\label{fig:add_full_sample}
\end{figure}

To mitigate the training--inference mismatch at large mask ratios, we further mix single-timestep full-image samples into each training batch. Specifically, for a fraction $\rho$ (in our experiments, we set  $\rho=0.25$) of samples in a mini-batch, we disable the dual-timestep and Attention Separation and assign all tokens the same timestep, i.e., $\tau_i=t$ for all $i$, while using the standard full-attention. The remaining samples are trained with the original dual-timestep scheduling and Attention Separation. Therefore, each batch contains both separated dual-timestep samples, which preserve the heterogeneous-noise and part-level data augmentation effects, and full-image single-timestep samples, which expose the model to the same global attention pattern used at inference. This mixed setting preserves the augmentation effect from dual-timestep scheduling and Attention Separation for a subset of samples, while also exposing the model to the standard full-image, single-timestep attention pattern used at inference. As shown in Figure~\ref{fig:add_full_sample}, this strategy substantially improves FID when the mask ratio is large. At $\alpha=0.50$, where all-sample Attention Separation gives each attention component only half of the image context, adding full-image samples reduces the 800K FID from $38.19$ to $24.15$. The gain becomes smaller as the mask ratio decreases. This trend is consistent with our interpretation: when $\alpha=0.25$, one token group already covers most of the image, so the separated training samples remain relatively close to the full-image inference condition. In this case, replacing part of the batch with vanilla single-timestep samples is less critical and may also weaken the augmentation effect, since those samples no longer receive either dual-timestep noise augmentation or Attention Separation.

\begin{table*}[t]
\centering
\captionsetup{width=0.7\textwidth}
\caption{\textbf{Quantitative results on ImageNet 256$\times$256} with Classifier-free Guidance (CFG)~\cite{cfg}. The \textbf{best} and \underline{second-best} results on each metric are highlighted in bold and underlined.}
\label{tab:imagenet_results_256}

\begin{tabular}{lcccccc}
\toprule
Model & Training Steps & FID$\downarrow$ & sFID$\downarrow$ & IS$\uparrow$ & Pre.$\uparrow$ & Rec.$\uparrow$ \\
\midrule
\multicolumn{7}{l}{\textit{vanilla diffusion transformers}} \\
DiT-XL/2   & 7M  &    2.27 & 4.60 & {278.2} & \textbf{0.83} & 0.57  \\
SiT-XL/2 & 7M & 2.06 & \textbf{4.50} & 270.3 & \underline{0.82} & 0.59 \\
\midrule
\multicolumn{7}{l}{\textit{representation alignment with external encoder}} \\
SiT-XL/2 +  REPA & 4M & \textbf{1.42} & 4.70 & 305.7 & 0.80 & \textbf{0.65} \\
\midrule
\multicolumn{7}{l}{\textit{self-representation alignment}} \\
SiT-XL/2 +  SRA & 4M & 1.58 & {4.65} &\underline{311.4} & {0.80} & \underline{0.63} \\
SiT-XL/2 +  Self-Flow & 4M & {1.47} & \underline{4.54} & {305.4} & {0.80} & 0.61 \\
\textbf{SiT-XL/2 +  ours} & 4M & \underline{1.44} & {4.60} & \textbf{315.3} & {0.81} & \underline{0.63} \\
\bottomrule
\end{tabular}
\end{table*}

\section{Putting Things Together}
\label{sec:putting_together}

The analyses above lead to a unified interpretation of the transition from SRA to Self-Flow. The key component that improves Self-Flow over SRA, dual-timestep scheduling, is not mainly explained by stronger self-supervision as suggested in the Self-Flow paper. By applying Attention Separation, we remove  token interactions of different noise-level while preserving the same heterogeneous-noise input, yet the performance does not degrade and can even improve. This indicates that the benefit of dual-timestep scheduling mainly comes from data augmentation that expands the effective training data: the same image is observed under more diverse noise states. We further find that Attention Separation itself also acts as an augmentation mechanism: by splitting one training image into multiple independently optimized token groups under shared model parameters, it increases the number of effective training views derived from the same sample. In this sense, the answer to the question in our title is that \textbf{the gain from SRA to Self-Flow is better understood as \emph{data augmentation}, rather than as stronger self-supervision.}

This interpretation naturally leads to our final training scheme. We retain the internal self-alignment objective of SRA, since it provides the representation-learning signal without relying on external encoders. On top of it, we use dual-timestep scheduling to augment each image along the noise-state dimension, and apply Attention Separation to further create part-conditioned training views. Both components are therefore used as augmentation mechanisms within the self-representation alignment framework for training.

\section{System-Level Comparison}

\subsection{Setup}
\noindent{\textbf{Implementation details.}}
Unless specified otherwise, our training pipeline closely mirrors the configurations established in privious baselines~\cite{dit,sit,sra,self-flow}. Specifically, we employ the AdamW optimizer~\cite{adamw} with a constant learning rate of 1e-4, zero weight decay, and a total batch size of 256, and uniform timestep sampling strategy. Latent representations are extracted utilizing the pre-trained Stable Diffusion VAE~\cite{sd-vae}. For the model backbone, we adopt the XL/2 SiT, all of which operate with a patch size of 2. For our method, we follow the setups of Self-Flow~\cite{self-flow} and SRA~\cite{sra}, where the alignment layer for the student and teacher are 8 and 20, respectively. The teacher is obtained via the Exponential Moving Average (EMA) of the student with a decay of 0.9999, and the coefficient of the alignment loss is set to 0.5. The mask ratio is set to 0.25 as it yields the best performance (ablated in Table~\ref{tab:mask_ratio_800k} and Figure~\ref{fig:add_full_sample}). All experiments are conducted on 8 NIVIDA H20 GPUs.

\noindent{\textbf{Evaluation metrics.}}
To evaluate generation quality, we report Fr\'echet Inception Distance (FID~\cite{fid}), sFID~\cite{sfid}, Inception Score (IS~\cite{IS}), along with precision and recall~\cite{pre-rec}. To ensure equitable comparisons with existing baselines, we compute these metrics using the official TensorFlow evaluation suite from ADM~\cite{adm} with 50K generated samples and the standard reference statistics.

\noindent{\textbf{Baselines for comparison.}}
We benchmark our method against vanilla DiT and SiT~\cite{dit,sit} as well as paradigms from both branches of representation alignment: namely, those with and without dependency on external models. Within each category, we benchmark against the representative method. Specifically, we select REPA~\cite{repa} as the representative for external-model-assisted alignment, and SRA~\cite{sra} and Self-Flow~\cite{self-flow} for self-alignment approaches.

\begin{table*}[t]
\centering
\captionsetup{width=0.7\textwidth}
\caption{\textbf{Quantitative results on ImageNet 512$\times$512} with Classifier-free Guidance (CFG)~\cite{cfg}.}
\label{tab:imagenet_results_512}
\begin{tabular}{lcccccc}
\toprule
Model & Training Steps & FID$\downarrow$ & sFID$\downarrow$ & IS$\uparrow$ & Pre.$\uparrow$ & Rec.$\uparrow$ \\
\midrule
\multicolumn{7}{l}{\textit{vanilla diffusion transformers}} \\
DiT-XL/2   & 3M  &    3.04 & 5.02 & {240.8} & \textbf{0.84} & 0.54  \\
SiT-XL/2 & 3M & 2.62 & 4.18 & 252.2 & \textbf{0.84} & 0.57 \\
\midrule
\multicolumn{7}{l}{\textit{representation alignment with external encoder}} \\
SiT-XL/2 +  REPA & 1M & \textbf{2.08} & 4.19 & 274.6 & \underline{0.83} & \underline{0.58} \\
\midrule
\multicolumn{7}{l}{\textit{self-representation alignment}} \\
SiT-XL/2 +  SRA & 1M & 2.17 & 4.15 & 279.3 & \underline{0.83} & \textbf{0.59} \\
SiT-XL/2 +  Self-Flow & 1M & \underline{2.12} & \textbf{4.10} & \underline{280.2} & \underline{0.83} & \underline{0.58} \\
\textbf{SiT-XL/2 +  ours} & 1M & \textbf{2.08} & \underline{4.12} & \textbf{282.7} & \underline{0.83} & \underline{0.58} \\
\bottomrule
\end{tabular}
\end{table*}

\begin{figure*}[t]
    \centering
    \vspace{1em}
    \includegraphics[width=1\linewidth]{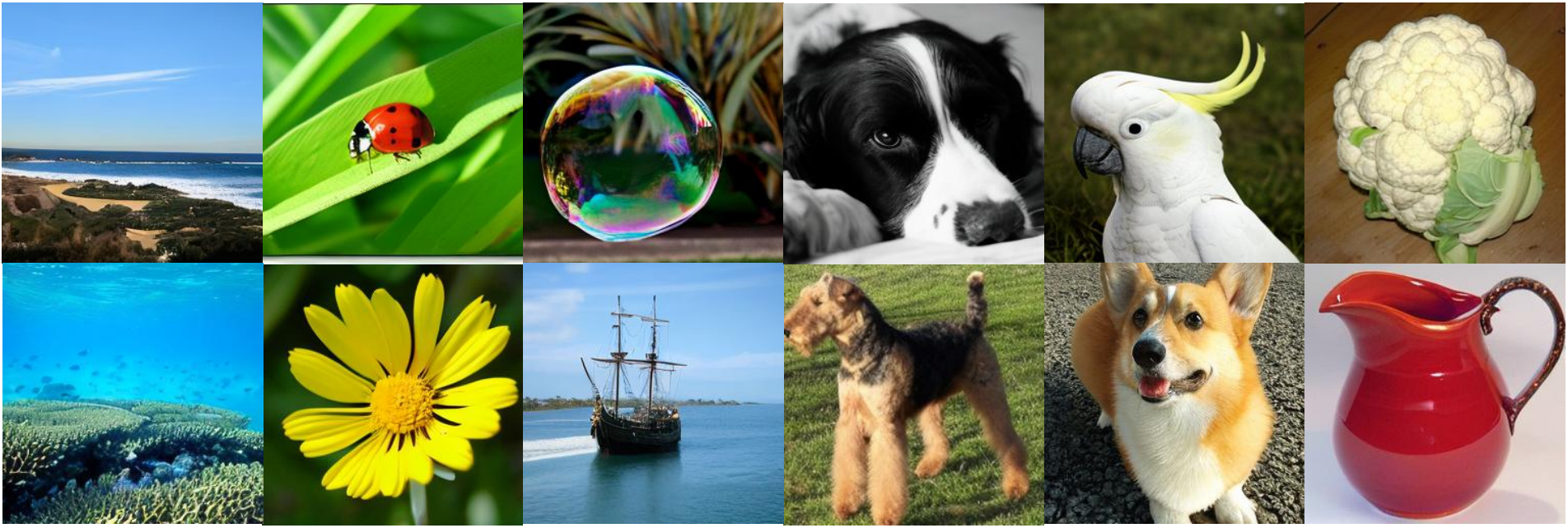}
    \caption{Qualitative results on ImageNet using SiT-XL + ours. We use classifier-free guidance with $w$ = 4.0.}
    \label{fig:selected_sample}
\end{figure*}

\subsection{Results}

\noindent{\textbf{Our method is competitive with both previous external and self-alignment methods.}}
Table~\ref{tab:imagenet_results_256} reports ImageNet $256\times256$ results. Compared with the vanilla SiT-XL/2 trained for 7M steps, our method reaches a lower FID using 4M steps, improving FID from $2.06$ to $1.44$ and IS from $270.3$ to $315.3$. Among self-alignment methods, our method improves over SRA and Self-Flow in FID and IS, achieving the best IS and the second-best FID among all compared methods. Although REPA obtains a slightly lower FID with an external pretrained encoder, our method remains comparable while relying only on self-representation alignment inside the diffusion transformer.

\noindent{\textbf{The same trend holds at higher resolution.}}
Table~\ref{tab:imagenet_results_512} shows the results on ImageNet $512\times512$. Our method matches the best FID of REPA at $2.08$, outperforms both SRA and Self-Flow in FID and IS, and achieves the highest IS of $282.7$. It also substantially improves over the vanilla SiT-XL/2 baseline trained for 3M steps, reducing FID from $2.62$ to $2.08$ with only 1M training steps. These results indicate that the augmentation interpretation developed in the controlled studies translates to stronger system-level performance, and that the resulting method remains effective when scaling to higher image resolution.

\section{Conclusion}
\label{sec:conclusion}

In this work, we revisit the transition from SRA to Self-Flow and study whether the improvement actually comes from. By introducing Attention Separation, we preserve the same heterogeneous-noise input while removing cross-noise token interaction. The resulting performance does not degrade and can even improve, indicating that the benefit of dual-timestep scheduling is better explained as noise-state data augmentation rather than cleaner-to-noisier token interaction alone. We further show that Attention Separation itself provides a part-level augmentation effect by splitting a single image into multiple effective training parts to expand the training data. Based on these findings, we combine dual-timestep scheduling and Attention Separation within the self-representation alignment framework. Experiments on ImageNet $256\times256$ and $512\times512$ show that this augmentation-based interpretation leads to a simple and effective training scheme, competitive with both external-encoder alignment and previous self-alignment methods.

{
    \small
    \bibliographystyle{ieeenat_fullname}
    \bibliography{main}
}

\end{document}